\documentclass[conference]{IEEEtran}
\IEEEoverridecommandlockouts
\usepackage{amsmath,epsfig}
\usepackage{algorithm}
\usepackage{algorithmic}
\usepackage{times}
\usepackage{helvet}
\usepackage{courier}
\usepackage{epsfig}
\usepackage{color}
\usepackage{xcolor}
\usepackage{colortbl}
\usepackage{tabularx}
\usepackage{gensymb}
\usepackage{amsfonts}
\usepackage{arydshln}
\usepackage{subcaption}
\usepackage{booktabs}
\usepackage{multirow}

\let\OLDthebibliography\thebibliography
\renewcommand\thebibliography[1]{
  \OLDthebibliography{#1}
  \setlength{\parskip}{0pt}
  \setlength{\itemsep}{0pt plus 0.3ex}
}

\begin{document}\sloppy

\def\x{{\mathbf x}}
\def\L{{\cal L}}

\title{PSPU: Enhanced Positive and Unlabeled Learning by Leveraging Pseudo Supervision}
%
\author{\IEEEauthorblockN{Chengjie Wang\textsuperscript{1,2}, Chengming Xu\textsuperscript{2}, Zhenye Gan\textsuperscript{2}, Yuxi Li\textsuperscript{2}, Jianlong Hu\textsuperscript{2,3}, 
Wenbing Zhu\textsuperscript{4,5}, Lizhuang Ma\textsuperscript{1,*}\thanks{*Corresponding authors.}}
\IEEEauthorblockA{\textit{\textsuperscript{1}{Shanghai Jiao Tong University}, \textsuperscript{2}{Tencent Youtu Lab}, \textsuperscript{3}Xiameng University, \textsuperscript{4}Fudan Univesity}, \textsuperscript{5}{Rongcheer Co., Ltd}\\
\{jasoncjwang,chengmingxu,wingzygan,yukiyxli\}@tencent.com, hujianlong@stu.xmu.edu.cn, \\
louis.zhu@rongcheer.com, ma-lz@cs.sjtu.edu.cn
}
}

\maketitle

\begin{abstract}
Positive and Unlabeled (PU) learning, a binary classification model trained with only positive and unlabeled data, generally suffers from overfitted risk estimation due to inconsistent data distributions. To address this, we introduce a pseudo-supervised PU learning framework (PSPU), in which we train the PU model first, use it to gather confident samples for the pseudo supervision, and then apply these supervision to correct the PU model's weights by leveraging non-PU objectives. We also incorporate an additional consistency loss to mitigate noisy sample effects. Our PSPU outperforms recent PU learning methods significantly on MNIST, CIFAR-10, CIFAR-100 in both balanced and imbalanced settings, and enjoys competitive performance on MVTecAD for industrial anomaly detection.
\end{abstract}
\section{Introduction}
\label{sec:intro}

Positive and Unlabeled (PU) learning is a binary classification task with only partial positive data annotated.
Such task is widely applicable in different real-life domains, e.g., fraud recognition in financial fields~\cite{chen2023precision}, fake detection in recommendation system~\cite{de2022network}, pathologic diagnosis in medical image processing, anomaly detection in industry, satellite image recognition, etc.
In these scenarios, some positive samples are well maintained, while plenty of unlabeled data are generated every day, and is unrealistic to be labeled manually and confidently. Moreover, in industry anomaly detection, most of the unlabeled data should be qualified, i.e. negative samples, to satisfy the requirement of manufacture. Similarly, most human beings to be diagnosed are healthy and negative for a specific disease. Such situation endows the imbalanced PU learning, where positive and negative samples suffer from large proportion gap in the training set, more practical value.


\begin{figure}[h]
  \centering
   \includegraphics[width=0.85\linewidth]{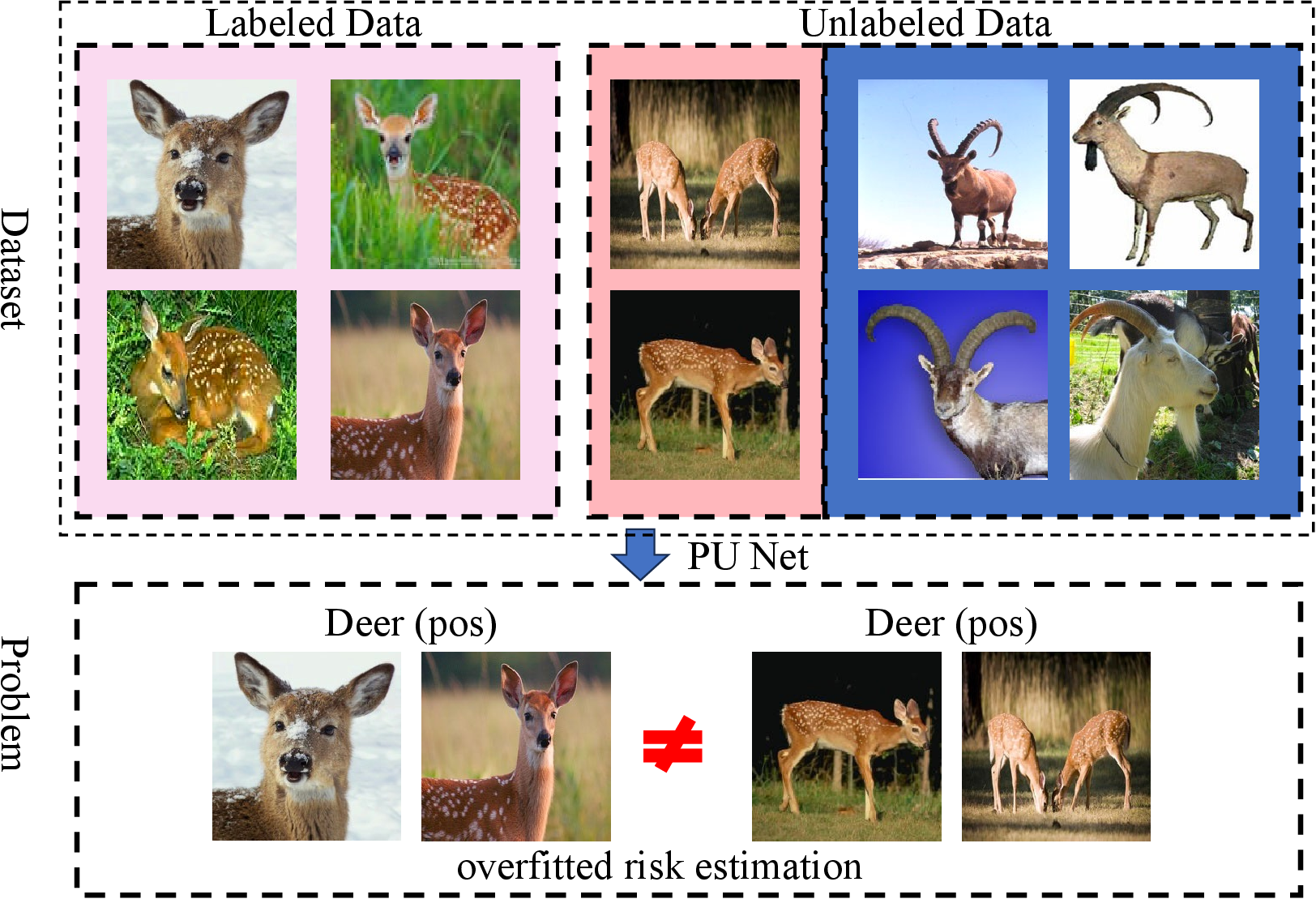}
    \caption{Challenges in PU net: traditional PU net suffers from overfitted risk estimation between labeled and unlabeled positive data. Note that images of \textcolor{red}{deer} denote the positive samples, and that of \textcolor{blue}{goat} denote the negative samples.}\label{problems_in_pu_and_ssl}
\end{figure}

Previous studies~\cite{du2015convex, niu2016theoretical, du2014analysis} enlighten that the core of PU learning is to extract reliable negative distribution information from unlabeled data while avoiding false negative. To achieve this goal, some advanced risk estimators with different constraints are proposed and properly integrated into model training, enabling the classifier to recognize the negative data. For example, unbiased PU learning (uPU)~\cite{du2014analysis} and non-negative PU leanring (nnPU)~\cite{kiryo2017positive} utilize a prior positive ratio to constrain the learning process on the unlabeled data.

The promising performance of these methods on naive and balanced vision datasets can be largely attributed to a strong \textit{selected completely at random} (SCAR) assumption that implies \emph{the distributions of both labeled and unlabeled positive data are similar}, which facilitates the usage the risk estimation of labeled positive data to describe that of the positive samples in unlabeled data.
However, we empirically find the assumption cannot be strictly satisfied, due to the random nature of training data sampling.
For example, as shown in Fig.~\ref{problems_in_pu_and_ssl}, the traditional PU net can fit labeled deer images (positive samples) well, but cannot cover the gap between risk of labeled data and that of unlabeled data (e.g., different numbers, and different posture), which leads to misclassification, especially when the number of positive samples is limited and imbalanced against negative samples. This problem, termed as \emph{overfitted risk estimation}, is mainly related to overfitting of labeled positive data in risk estimation. 

To solve this problem, in this paper, we propose a novel PU learning framework named Pseudo-Supervised PU learning (PSPU). In PSPU, we take advantage of the previous PU risk estimator to generate pseudo supervision, which is then involved in the proposed pseudo-supervised training adopting by actively adopting various objective functions. Such a pipeline can explicitly extracts and fits positive samples from unlabeled data and is less affected by the overfitted risk estimations in labeled positive data. Therefore, the learned parameters can effectively correct any misguidance in PU learning. Moreover, a feature-based self-supervised consistency loss is integrated into the pseudo-supervised training, to reduce the negative impacts from the noisy samples. In order to further prevent the overfitted risk estimation from the basic PU net, we design a weights transfer operation between the basic PU net and pseudo-supervised net. 

The excellent performance of the proposed PSPU is comprehensively demonstrated on both balanced and imbalanced PU learning tasks among normal vision data including MNIST, CIFAR-10 and CIFAR-100 and industrial data such as MVTecAD. Our main contributions are summarized as follows:

\noindent1. A practical framework PSPU is proposed for PU learning, which leverages the cooperation of basic PU net with pseudo-supervised training to {mitigate} the overfitted risk estimation problem in traditional PU learning methods.

\noindent2. {A reliable pseudo supervision generation process guided by PU learning is designed to alleviate mislabeled data.}

\noindent3. {The customized pseudo-supervised training framework is incorporated into PSPU via a special progressive knowledge transfer to feedback knowledge and help PU training.}

\section{Related Work}

\subsection{Positive-Unlabeled Learning}\label{related_PU_learning}
Positive-Unlabeled Learning (PU Learning) is a variant of the classical binary classification problem where the training dataset consists of only a set of positive samples and a set of unlabeled samples. Research related to PU learning appeared in the 2000s and has gradually attracted attention in recent years because PU Learning has applicability in many real-world problems. Among the main research categories of PU learning~\cite{bekker2020learning, wang2021asymmetric}, biased learning treats the set of unlabeled samples as negatives with noisy labels~\cite{claesen2015a}. Two-step approaches~\cite{he2020instancede, xu2022split} focus on identifying reliable negative set and further training. As for the imbalanced PU learning, Sakai et al.~\cite{sakai2018semi} developed a pseudo-supervised AUC optimization method to deal with imbalanced problem. Su et al.~\cite{su2021positive} proposed a novel reweighting strategy for PU learning to deal with the problem of imbalanced data. Zhao et al.~\cite{zhao2022dist} proposed to solve PU learning via aligning label distribution. However, the available data in many real-world scenarios is extremely imbalanced, such as medical diagnosis and recommendation systems, and such data may aggravate the overfitted risk estimation problem. In contrast, our proposed method can effectively solve such problems, leading to more reliable performance.

\subsection{Semi-supervised Learning}
Semi-supervised learning (SSL) is a learning paradigm that leverages a large amount of unlabeled data to improve the performance of a learning algorithm along with a small amount of labeled data~\cite{yang2021aso}. The concept of semi-supervised learning first appeared in the 1970s. Currently there are various methods of semi-supervised learning. For example, Graph-based methods, such as DNGR and GraphSAGE~\cite{hamilton2017inductive}, use graph embedding to extract a graph from the training dataset where nodes denote training samples and edges represent similarity measurement of node pairs.
Pseudo-labeling methods, such as EnAET~\cite{wang2021enaet} and Semihd~\cite{imani2019semihd}, leverage unlabeled data by generating semi-labels with high confidence and add them to the training dataset to improve the performance of the model. In this paper, we explore the possibility of engaging semi-supervised objectives in training of PU learning models. 

\section{Methodology}\label{sec:method}

\subsection{Preliminary}
\noindent\textbf{Problem Formulation.} PU learning can be seen as a special case of binary classification. Given a training set $\mathcal{D}^{train}$ which contains positive data $\mathcal{D}^p=\{x^p_i, y^p_i\}_{i=1}^{N_p}$ and unlabeled data $\mathcal{D}^u=\{x^u_i, y^u_i\}_{i=1}^{N_u}$, where $x\in\mathbb{R}^{3\times H\times W}$ denotes input images, $y\in\{\pm{1}\}$ denotes a binary label, $N_p, N_u$ denote number of positive and unlabeled samples, the goal is to learn a binary classifier $g$ with strong inference power on test data $\mathcal{D}^{test}$. Additionally, in PU learning, the prior probability of the positive samples $\pi_p=\mathcal{P}(y=+1)$ is assumed to be available, and $\pi_n=1-\pi_p$ is prior for negative data.

\noindent\textbf{PU Learning with Empirical Risk Estimator.}
Basically, the empirical risk for a binary classification is composed of positive risk $\mathcal{R}_p^+(g)$ and negative risk $\mathcal{R}_n^{-}(g)$. Generally, $\mathcal{R}_p^+(g)$ can be empirically estimated as $\hat{\mathcal{R}}_p^+(g)=\frac{1}{N^p}\sum_{x_i\in \mathcal{D}^p}l(g(x_i),+1)$, where $l$ denotes a binary classification objective function. Since negative labeled data is not available in PU learning, $ \mathcal{R}_{n}^{-}(g)$ can not be directly estimated. As a solution, the previous methods such as uPU and nnPU take a simple transformation:
\begin{align}
    \mathcal{R}^{-}(g)  &= \mathbb{E}[l(g(x),-1)] \nonumber \\
        & = \pi_p\mathbb{E}[l(g(x),-1)|y=+1] + \nonumber\\
        & \quad \quad \pi_n\mathbb{E}[l(g(x),-1)|y=-1]  \nonumber\\
        & = \pi_p\mathcal{R}_{p}^{-}(g) + \pi_n\mathcal{R}_{n}^{-}(g).
\end{align}
In this way $\pi_n\mathcal{R}_{n}^{-}(g)$ can be alternatively calculated as 
\begin{equation}\label{equation_rn}
    {\pi_n\mathcal{R}_{n}^{-}(g)}={{\mathcal{R}}^{-}(g)}- \pi_p{{\mathcal{R}}_{p}^{-}(g)}.
\end{equation}
in which ${\mathcal{R}}^{-}(g)$ and ${\mathcal{R}}_{p}^{-}(g)$ can be empirically estimated from unlabelled and positive data respectively. Based on such an idea, these methods can train a binary classifier without access of negative labelled data using $\mathcal{L}_{pu}=\pi_p\mathcal{R}_{p}^{+}(g) + \mathcal{R}_{n}^{-}(g) - \pi_p\mathcal{R}_{p}^{-}(g)$.

\subsection{Motivation}\label{motivation}

\begin{figure}
  \centering
  \begin{subfigure}{0.47\linewidth}
    \includegraphics[width=1.0\columnwidth]{{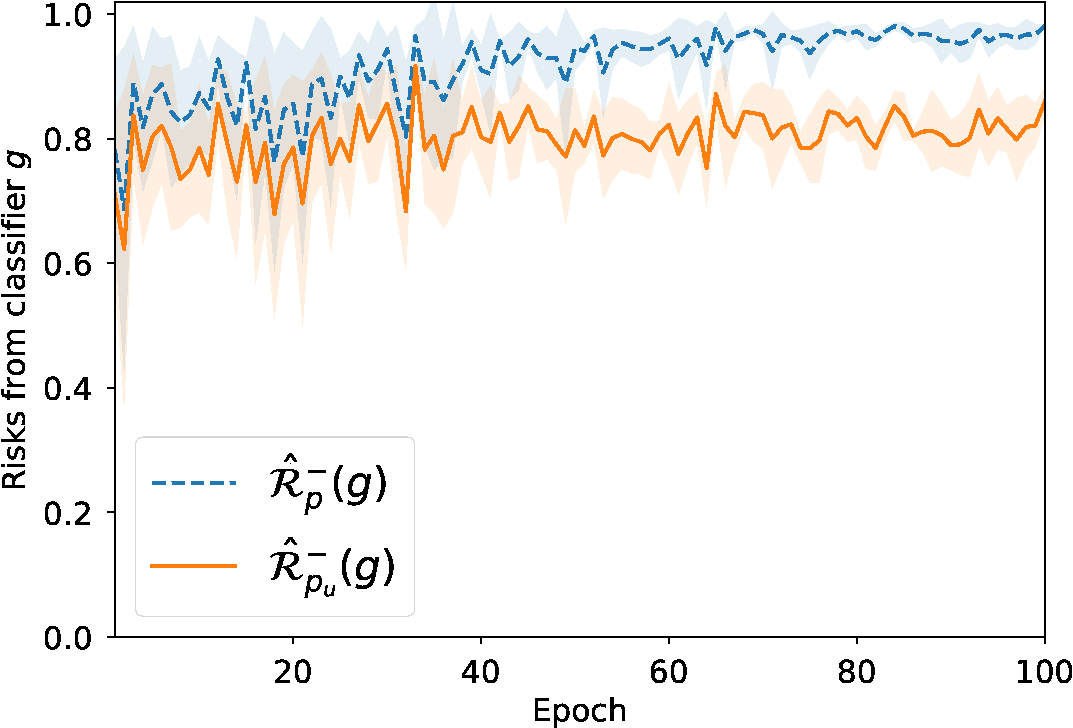}}
    \caption{$\pi_p=0.4$}
    \label{fig_positive_loss_b}
  \end{subfigure}
  \hfill
  \begin{subfigure}{0.47\linewidth}
    \includegraphics[width=1.0\columnwidth]{{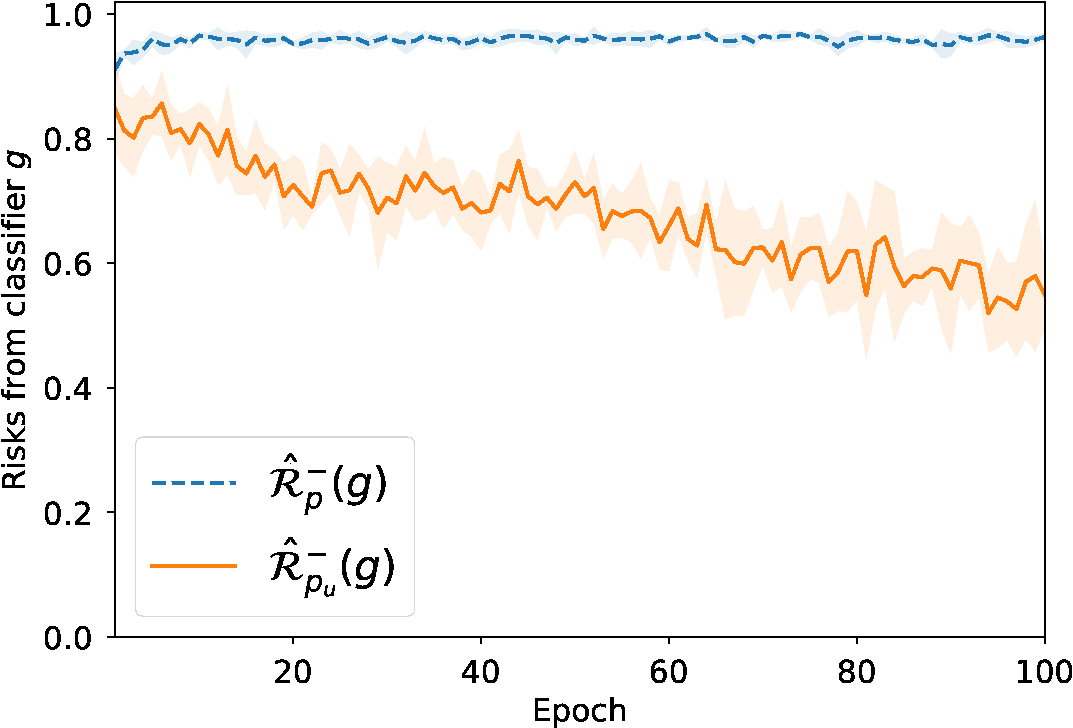}}
    \caption{$\pi_p=0.1$}
    \label{fig_positive_loss_c}
  \end{subfigure}
  \caption{Gap between PU risk estimation and oracle risk estimation among different settings.}
  \label{fig_positive_loss}
  \vspace{-0.2in}
\end{figure}

One main problem of Eq~(\ref{equation_rn}) is that the estimation of $\mathcal{R}^{-}(g)$ and $\mathcal{R}_p^{-}(g)$ individually based on $\mathcal{D}^u$ and $\mathcal{D}^p$  strongly relies on the assumption that the distribution $\mathcal{D}^p$ should be identical to that of the unlabeled positive data covered by $\mathcal{D}_u$. However, as illustrated in Sec.~\ref{sec:intro}, PU learning problems often raise in the scenario of label scarcity. For example, in industrial production, it is hard to collect sufficient labeled data for various defects. In this way, $\mathcal{D}^p$ can be severely biased from $\mathcal{D}^u$. Consequently, the approximation of ${\mathcal{R}}_{p}^{-}(g)$ can be inaccurate, thus breaking the assumption and making the approximation of ${\mathcal{R}}_{p}^{-}(g)$ unfavourable. Such a problem can be more harmful as the size of $\mathcal{D}^p$ gets smaller and the whole training set gets more imbalanced.

To further clarify this problem, in Fig.~\ref{fig_positive_loss}, we analyze two statistically estimated risks from classifier $g$ on CIFAR-10: one is risk $\hat{\mathcal{R}}_{p}^{-}(g)$ on positive data from $\mathcal{D}^p$, the other is oracle risk $\hat{\mathcal{R}}_{pu}^{-}(g)$ calculated from $\mathcal{D}^u$ in a similar way as $\mathcal{D}^p$, assuming the labels are available. We compare the difference between these two statistics among different choices of positive prior $\pi_p$. It is easy to observe that the gap between $\hat{\mathcal{R}}_{p}^{-}(g)$ and $\hat{\mathcal{R}}_{pu}^{-}(g)$ in Fig.\ref{fig_positive_loss}(a) and Fig.\ref{fig_positive_loss}(b) becomes larger as the training goes on, indicating more overfitted risk estimation and thus empirically proving our claim. 

One would ask if it is possible to directly utilize other objective functions to avoid estimating $\mathcal{R}_{p}^{-}(g)$ from positive and unlabeled data. Unfortunately, it is non-trivial to adopt commonly-used objectives such as those in fully-supervised learning or pseudo-supervised training, due to the absence of negative labeled data. To this end, we present PSPU, a two-stage framework that can better handle the PU learning problem especially for imbalanced data.

\subsection{PSPU Framework}

\begin{figure}[H]
  \centering
   \includegraphics[width=0.9\linewidth]{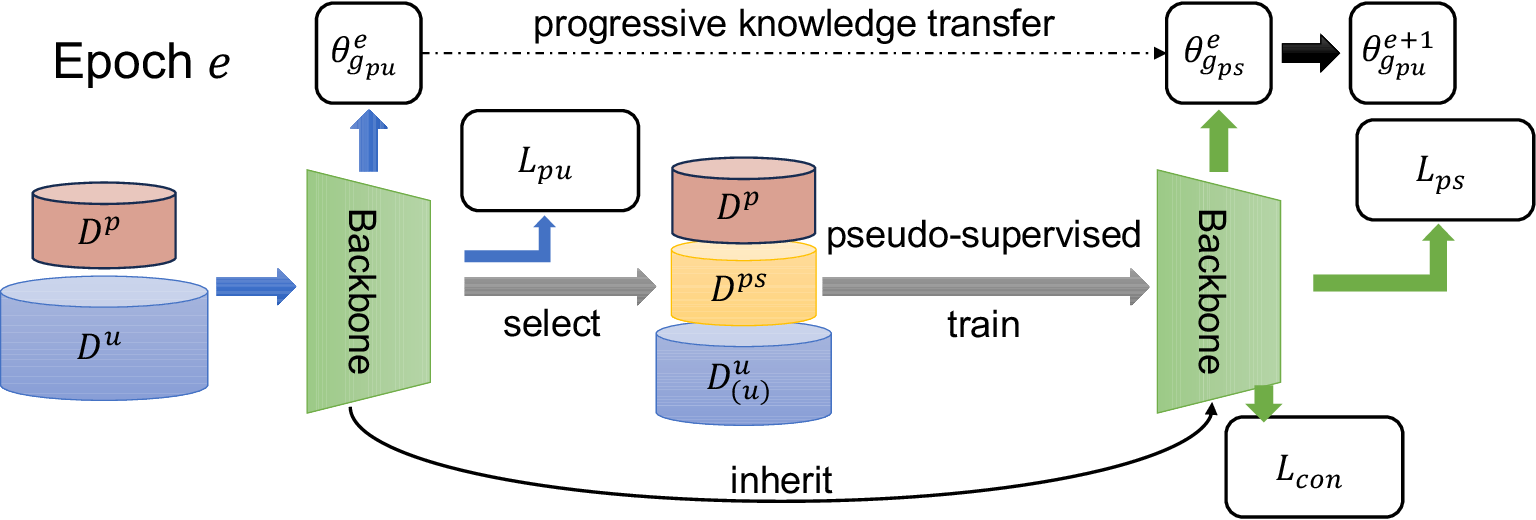}
   \caption{The Training Process of PSPU at Epoch $e$.}\label{fig_PSPU_framework}
\end{figure}

As shown in Fig.~\ref{fig_PSPU_framework}, our method iteratively utilizes two stages, namely pseudo supervision generation and pseudo-supervised training. The model first generates smooth pseudo supervision from unlabeled data by leveraging traditional PU learning methods. Then these supervision is used along with the positive data and rest unlabeled data to train another model without using PU risk estimators. Through dense knowledge transfer between models supervised by these two stages, our method can generally avoid the drawbacks of the PU risk estimators, thus leading to more desirable performance.

\noindent\textbf{Pseudo Supervision Generation.} In each epoch, we first train a basic PU net following nnPU~\cite{kiryo2017positive} and obtain a PU classifier $g_{pu}$, which can provide the pseudo supervision for further training. While most previous methods directly utilize the pseudo labels predicted by $g_{pu}$, the estimation bias as mentioned in Sec.~\ref{motivation} can lead to unreliable pseudo labels, thus misguiding the model. To solve this problem, we propose a supervision mixture mechanism based on Mixup. Specifically, from $\mathcal{D}^u$, we first select $n_s$ most confident positive and negative predictions by $g_{pu}$ respectively. Accordingly, $\mathcal{D}^u$ can be divided into three distinctive part as $\mathcal{D}^u=\mathcal{D}^u_{(p)}\cup \mathcal{D}^u_{(n)}\cup \mathcal{D}^u_{(u)}$, where $\mathcal{D}^u_{(p)}$ and $\mathcal{D}^u_{(n)}$ denote samples with confident pseudo positive and negative labels. $\mathcal{D}^u_{(u)}$ is the set of remained unlabeled data. Then samples from $D^{u}_{(n)}$ and $D^{u}_{(p)}$ are mixed to form the psuedo supervision set $\mathcal{D}^{ps}$. In particular, for all $(x_i, 1), (x_j, -1) \in \mathcal{D}^u_{(p)}\cup \mathcal{D}^u_{(n)}$ we have $x'=\beta x_i + (1-\beta)x_j, y'=\beta y_i + (1-\beta)y_j,$ where $\beta$ is randomly sampled from Beta distribution. Note that different from Mixup, in which ground truth samples are mixed, we leverage such technique to interpolate pseudo-labeled samples to further alleviate the effect from mislabeled samples. Consequently, we construct a new dataset $\mathcal{D}'=\{\mathcal{D}^p, \mathcal{D}^{ps}, \mathcal{D}^u_{(u)}\}$ for the following pseudo-supervised training.

\noindent\textbf{Pseudo-supervised Training.}\label{semi_supervised_training}
The $\mathcal{D}'$ constructed in the above step can help us build a new training objective without using PU risk estimator. We find that such new pseudo supervision can well facilitate the commonly-used semi-supervised learning (SSL) objectives, with the negative labeled samples replaced with the samples in $\mathcal{D}^{ps}$. Formally, a new network $g_{ps}$ is initialized based on $g_{pu}$ and optimised with a versatile pseudo-supervised loss function, which can be instantiated as commonly-used SSL methods such as MixMatch~\cite{berthelot2019mixmatch}, FixMatch~\cite{sohn2020fixmatch}, FlexMatch~\cite{zhang2021flexmatch}, etc. Such non-PU objectives, denoted as $\mathcal{L}_{ps}$, do not depend on how close the distribution $\mathcal{D}^p$ and the real positive distribution are, hence can better handle the previously explained problems of biased risk, thus better correcting the model weights. Moreover, different from directly applying pseudo-supervised to raw or pseudo-labeled unlabeled data, which is adopted in the previous works, utilizing the pseudo supervision as introduced above can help the model alleviate the problem of imbalanced and noisy supervision, thus leading to better performance especially for the imbalanced training data.

To further enhance the robustness against noise, we introduce an additional feature consistency loss $\mathcal{L}_{con}$. To be specific, for two augmented view $x_{aug1}, x_{aug2}$ of the same data $x$, we extract their features $\delta_{x_{aug1}}, \delta_{x_{aug2}}$ after all blocks from the backbone network, and minimize their KL divergence as $\mathcal{L}_{con}(f)=D_{KL}(\delta_{x_{aug1}}||\delta_{x_{aug2}})$.

\noindent\textbf{Progressive Knowledge Transfer.}\label{classifier_weights_transfer}
After the steps above, we can obtain two different binary classifiers: $g_{pu}$ and $g_{ps}$ with the same backbone but stemming from PU training and pseudo-supervised training respectively. A straightforward strategy is to inject the weight of $g_{ps}$ to $g_{pu}$ for pseudo supervision generation in the next epoch. However, we empirically find that such operation may potentially lead to over-correction of the overfitted risk estimation. Therefore, we adopt a progressive knowledge transfer strategy. Specifically, after the $e$-th epoch, the initial weight for pseudo supervision generation is set as  $\theta_{g_{pu}}^{(e+1)}=\lambda\theta_{g_{pu}}^{(e)}+(1-\lambda)\theta_{g_{ps}}^{(e)}$, where $\theta$ denotes network parameters, $\lambda$ is the transfer co-efficient. In this way, knowledge learned from $g_{ps}$, which is benefited from pseudo-supervised training, can be effectively merged with the original capacity contained in $g_{pu}$ gradually through the iterative training. 

\begin{algorithm}[htp]
\caption{PSPU}\label{alg_sspu}
\small
\begin{algorithmic}[1]
\REQUIRE PU Data $\mathcal{D}=\mathcal{D}^p\bigcup \mathcal{D}^u$
\ENSURE Classifier $g_{pu}$, $g_{ps}$
\STATE e = 0
\WHILE{$e<STOP\_E$}
\STATE select batch $\mathcal{D}^{p,i} \subset \mathcal{D}^p$ and $\mathcal{D}^{u,i} \subset \mathcal{D}^u$
\STATE update $g_{pu}$ with ${\mathcal{L}_{nnPU}}(g_{pu}, \mathcal{D}^{p, i}, \mathcal{D}^{u, i})$
\STATE obtain classification scores $S_{u}(g, \mathcal{D}^{u})$ on $\mathcal{D}^{u}$ 
\STATE select $n_{s}$ negative samples by $S_{u}(g, \mathcal{D}^{u})$ as $\mathcal{D}^u_{(n)}$
\STATE select $n_{s}$ positive samples by $S_{u}(g, \mathcal{D}^{u})$ as $\mathcal{D}^{u}_{(p)}$
\STATE mix $\mathcal{D}^u_{(n)}$ and $\mathcal{D}^{u}_{(p)}$ to get $\mathcal{D}^{ps}$
\STATE construct $\mathcal{D}'=\{\mathcal{D}^p, \mathcal{D}^{ps}, \mathcal{D}^u_{(u)}\}$
\STATE select batch $\mathcal{D}^{p,j} \subset \mathcal{D}^p$ and $\mathcal{D}^{ps,j} \subset \mathcal{D}^{ps}$ $\mathcal{D}^{u,j}\subset \mathcal{D}^u_{(u)}$
\STATE update $g_{ps}$ using pseudo-supervised training on $\mathcal{D}'$
\STATE weights transfer $\theta_{g_{pu}}^{(e+1)}=\lambda\theta_{g_{pu}}^{(e)}+(1-\lambda)\theta_{g_{ps}}^{(e)}$
\STATE $e\leftarrow e + 1$
\ENDWHILE
\RETURN $g_{pu}$, $g_{ps}$
\end{algorithmic}
\end{algorithm}

\begin{figure}[t]
  \centering
  \begin{subfigure}{0.47\linewidth}
   \includegraphics[width=1.0\columnwidth]{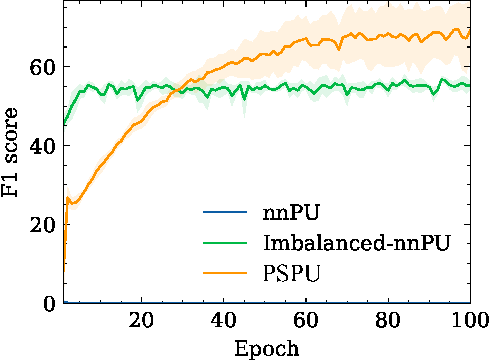}
   \caption{}
   \label{fig_imblanced_f1}
  \end{subfigure}
  \begin{subfigure}{0.47\linewidth}
   \includegraphics[width=1.0\columnwidth]{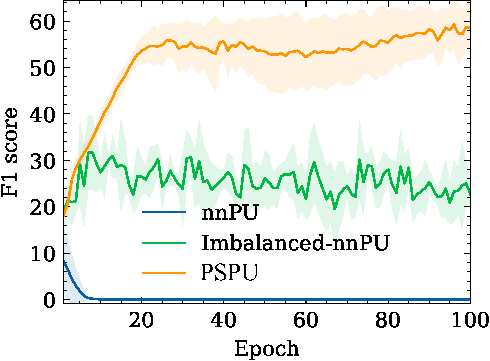}
   \caption{}
   \label{fig_extreme_imblanced_f1}
  \end{subfigure}
  \caption{Precision or F1 curve during training. (a) F1 score on imbalanced CIFAR-10. (b) F1 score on extremely imbalanced CIFAR-10.}
  \vspace{-0.1in}
\end{figure}

\section{Experiments}

\subsection{Comparison to State-of-the-Art Methods} 

To showcase the efficacy of our proposed PSPU, we conduct extensive experiments to compare our method with previous PU learning methods on multiple datasets. For imbalanced setting, CIFAR-10, CIFAR-100 and MVTecAD~\cite{bergmann2021mvtec} are adopted. For balanced setting, MNIST and CIFAR-10 are used, following previous works. Please refer to the supplementary material for detailed information and implementation detail of our method. 

\noindent\textbf{Imbalanced PU learning.} We first focus on imbalanced settings. 
Tab.~\ref{table_inbalanced_cifar10_0.1} and Tab.~\ref{table_MVTecAD} report the results on general imbalanced CIFAR-10 and MVTecAD, in which we adopt ReMixMatch~\cite{berthelot2019remixmatch} as the instantiation of pseudo-supervised training. The results show that compared with nnPU and ImbalancednnPU, PSPU achieves great improvement, with about $17\%$ F1 score improvement on CIFAR-10. This means PSPU enables to handle the situation with data imbalance. As for MVTecAD, our proposed PSPU enjoys 11\% F1 score advancement against HolisticPU and 10\% accuracy advancement against imbalanced nnPU. 


Moreover, the performance of extremely imbalanced PU learning on CIFAR-10 and CIFAR-100 is presented in Tab.~\ref{table_inbalanced_cifar10_0.01}, where part of positive samples are deleted from the unlabeled data to aggravate the imbalance. We test several implementations of pseudo-supervised training in this setting. We can find that the commonly-used nnPU and DistPU totally fail on CIFAR-10, and all of these competitors suffer from such a challenging setting. On the contrary, our proposed PSPU receives significantly better AUC and F1 score, indicating the effectiveness of our pipeline. Meanwhile, our method shows great versatility among different choice for the objective in pseudo-supervised training. We note that although HolisticPU performs better in terms of F1 score when comparing with PSPU-FlexMatch, it suffers from much lower accuracy and AUC, while our model has better balance between accuracy and F1 score, indicating better capacity when dealing with imbalanced data.

Fig.~\ref{fig_imblanced_f1} and Fig.~\ref{fig_extreme_imblanced_f1} show the F1 score of nnPU, Imbalanced-nnPU and PSPU during the training process on imbalanced CIFAR-10 and extremely imbalanced CIFAR-10. Note that the F1 score of nnPU drops rapidly to 0, this is due to the imbalance of the data lead the model to regard all inputs as negative samples. On the other hand, on extremely imbalanced data, the F1 scores of Imbalanced-nnPU fluctuate substantially and result in significantly lower results than the general imbalanced case. 
In contrast, PSPU shows a relatively smooth trend and converges to a much better performance than the other competitors, indicating PSPU is capable of handling extremely imbalanced situations.

\noindent\textbf{Balanced PU learning.} Tab.~\ref{table_balanced_data} gives the accuracy comparison on balanced MINST and CIFAR-10 dataset. Compared with uPU, nnPU, DAN, Self-PU and DistPU, in balanced data,  PSPU improves $3.94\%$ and $3.11\%$ of the accuracy over the best results on MINST and CIFAR-10 respectively, showing the generalization ability of our method.

\begin{table}[t]
    \centering
    \caption{Results on general imbalanced CIFAR-10}
    \vspace{-1mm}
    \resizebox{0.48\textwidth}{!}
    {
    \begin{tabular}{cccc}
        \toprule
        Method & ACC \% & F1 \% & AUC \% \\
        \midrule
        nnPU~\cite{kiryo2017positive} & 90.00$\pm$(0.0)& 0$\pm$(0.0) & 50.00$\pm$(0.0) \\
        Imblanced-nnPU~\cite{su2021positive} & 90.47$\pm$(0.01)& 55.84$\pm$(0.02) & 88.98$\pm$(0.01) \\
        DistPU~\cite{zhao2022dist} & 93.21$\pm$(0.22)&	66.57$\pm$(0.01) &	93.64$\pm$(0.12)  \\
        HolisticPU~\cite{wang2023beyond} & 91.41$\pm$(0.26)&	67.20$\pm$(0.15) &	96.31$\pm$(0.20) \\
        \hline
        \hline
        PSPU-ReMixMatch & \textbf{96.04$\pm$(0.19)}& \textbf{80.16$\pm$(0.07)} & \textbf{97.49$\pm$(0.09)} \\
        \bottomrule
    \end{tabular}
    }
    \label{table_inbalanced_cifar10_0.1}
    \vspace{-0.2in}
\end{table}

\begin{table}[htp]
    \centering
    \caption{Results on MVTecAD dataset, which reflects imbalanced PU learning in real-life scenarios.}
    \vspace{-1mm}
    \resizebox{0.4\textwidth}{!}
    {
        \begin{tabular}{cccc}
            \toprule
            Method	&	ACC \%& F1 \% & AUC \% \\
            \midrule
            Imbalanced nnPU~\cite{su2021positive}	& 56.59	& 49.57 	& 82.17 	\\
            DistPU~\cite{zhao2022dist}	&	53.66 & 	45.35	& 82.96   \\
            HolisticPU~\cite{wang2023beyond}	&	50.84 & 	52.19	& 71.98   \\
            \hline
            \hline
            PSPU-ReMixMatch	&	\textbf{67.45}&	\textbf{63.41}  	& \textbf{86.57} 		\\
            \bottomrule
        \end{tabular}
    }
    \label{table_MVTecAD}
    \vspace{-0.1in}
\end{table}

\begin{table}[htp]
    \centering
    \caption{Results on extremely imbalanced CIFAR-10 and CIFAR-100}
    \vspace{-1mm}
        \resizebox{0.48\textwidth}{!}
    {
    \begin{tabular}{c|ccc|ccc}
        \toprule
         & \multicolumn{3}{c}{CIFAR-10}  & \multicolumn{3}{c}{CIFAR-100}\\
        Method & ACC \%& F1 \% & AUC \% & ACC \% & F1 \% & AUC \% \\
        \midrule
        nnPU~\cite{kiryo2017positive} & 90.00& 0  & 50.00  & 99.00& 0 & 50.00   \\
        Imblanced-nnPU~\cite{su2021positive} & 90.48 & 22.75 & 76.24   & 98.09& 45.50 & 96.03  \\
        DistPU~\cite{zhao2022dist} & 90.00&	0 &	84.94  &	99.14&	36.77 &	97.01 \\
        HolisticPU~\cite{wang2023beyond} & 92.05&	43.66	 & 88.20  &	92.78&	55.98 &	91.68 \\
        
        \hline
        \hline
        PSPU-MixMatch  & \textbf{94.29}& \textbf{65.56}& 99.38 & 91.31& 72.32  &\textbf{98.26} \\
        PSPU-ReMixMatch& 93.24& 64.18 & \textbf{91.41} & \textbf{99.46 }& \textbf{75.29}& 96.65  \\
        PSPU-FlexMatch  & 92.53& 52.51  & 90.76 & 98.50& 50.98 & 98.01 \\
        \bottomrule
    \end{tabular}
}
    \label{table_inbalanced_cifar10_0.01}
    \vspace{-0.2in}
\end{table}

\begin{table}[H]
  \caption{The accuracies on balanced data. ``*'' indicates training with more labeled positive data}\label{table_balanced_data}
  \vspace{-1mm}
  \centering
    \resizebox{0.48\textwidth}{!}
    {$
        \begin{tabular}{cccc}
        \toprule
        Method & Positive Size & MNIST \% & CIFAR-10 \%\\
        \midrule
        uPU~\cite{du2014analysis}  & $n_p=1000$ & 92.52$\pm$(0.39) & 88.00$\pm$(0.62)\\
        nnPU~\cite{kiryo2017positive} & $n_p=1000$ & 93.41$\pm$(0.20) & 88.60$\pm$(0.40)\\
        VPU$^*$~\cite{chen2019variational} & $n_p=3000$ & - & 89.70$\pm$(0.40)\\
        Self-PU~\cite{chen2020self} & $n_p=1000$ & 94.21$\pm$(0.54) & 89.68$\pm$(0.22)\\
        Imblanced-nnPU~\cite{su2021positive} & $n_p=1000$ &- & 89.52$\pm$(0.67)\\
        DistPU~\cite{zhao2022dist} & $n_p=1000$ & 90.93$\pm$(0.22) & 91.56$\pm$(0.52)\\
        \hline
        \hline
        PSPU-ReMixMatch & $n_p=1000$ & \textbf{98.15$\pm$(0.16)} & \textbf{94.67$\pm$(0.36)}\\
        \bottomrule
        \end{tabular}
    $}
    \vspace{-0.1in}
\end{table}

\subsection{Ablation Study}

{In this section, we design ablation experiments on PSPU to investigate how the two components of this framework, psedo supervision generation, and pseudo-supervised training, affect the performance of imbalanced data. If not specified, all the experiments are conducted based on the division of extremely imbalanced CIFAR-10.}

\begin{table}[H]
  \caption{The results of naive data selection in PU net, and our proposed pseudo supervision generation.}
  \vspace{-1mm}
  \label{table_mixup}
  \centering
    \resizebox{0.45\textwidth}{!}
    {$
        \begin{tabular}{cccc}
        \toprule
        Setting & ACC \%& F1 \% & AUC \% \\
        \midrule
        Vanilla & 89.72$\pm$(3.04)&	58.52$\pm$(4.89) &	{89.99$\pm$(0.88)} \\
        Pseudo & \textbf{94.29$\pm$(0.71)}& \textbf{65.56$\pm$(3.98)} & \textbf{91.31$\pm$(1.02)} \\
        \bottomrule
        \end{tabular}
    $}
    
    \vspace{-0.2in}
\end{table}

\begin{table}[H]
  \caption{Results with different selection number $n_{s}$ on imbalanced Cifar-10}
  \vspace{-1mm}
  \label{results_of_np_selection}
  \centering
    \resizebox{0.45\textwidth}{!}
    {$
        \begin{tabular}{cccc}
        \toprule
        sampling number $n_s$ & ACC \%& F1 \% & AUC \% \\
        \midrule
        $n_s={0.1\pi_{p}{n_u}}$ & 89.08$\pm$(1.38)& 63.38$\pm$(2.88) & 96.85$\pm$(0.31) \\
        $n_s={0.5\pi_{p}{n_u}}$ & 96.50$\pm$(0.40)& 83.10$\pm$(1.55) & \textbf{97.48$\pm$(0.27)} \\
        $n_s={\pi_{p}{n_u}}$ & \textbf{96.81$\pm$(0.43)} & \textbf{84.31$\pm$(3.01)} & 97.43$\pm$(0.44)\\
        \bottomrule
        \end{tabular}
    $}
    \vspace{-0.1in}
\end{table}

\noindent\textbf{Effectiveness of pseudo supervision.} We evaluate how the pseudo supervision can help the overall performance by comparing two variants: (1) \textbf{Vanilla}: the selected data $D_{u}^{(n)}, D_{u}^{(p)}$ is directly fed to following pseudo-supervised training, (2) \textbf{Pseudo}: our proposed pseudo supervision generation. The results shown in Tab.~\ref{table_mixup} reflect that the vanilla methods is much worse than the our full method, indicating our superiority against the commonly-used pseudo-label strategy.

\noindent\textbf{How are the results affected by the data size $n_{s}$ selected by PU net?} In addition, we also discuss about the impact of selection number $n_p$ in pseudo supervision generation.
Since $\pi_{p}$ is assumed to be available, positive sample size in the unlabeled set $\mathcal{D}^{u}$ can be estimated as $\pi_{p}N_u$. We investigate the size of sampled data of a pre-defined sample ratio $r$, and $n_s=r\pi_p n_u$. We compare the results with three sampling ratios: $r=10\%$, $r=50\%$ and $r=100\%$, respectively.
The results on normal imbalanced Cifar-10 are shown in Tab.~\ref{results_of_np_selection}.
It shows that the performance even when we half the sampled data size ($r=50\%$), the results keeps similar to the counterpart with $r=100\%$.

\noindent\textbf{The importance of Progressive Knowledge Transfer.} The operation of progressive knowledge transfer is an important mechanism in PSPU to ensure the knowledge feedback from pseudo-supervised training framework $g_{ps}$ to PU net $g_{pu}$. To analyze this operation quantitatively, we compare different style of cooperation between the weights of classifier $g_{pu}$ and $g_{ps}$. Concretely, we evaluate three different manners of information exchange between these two networks: (1) classifier $g_{pu}$ and $g_{pu}$ are optimized independently ($g_{pu}\neq g_{ps}$), (2) classifier $g_{ps}$ inherits weights of $g_{pu}$ ($g_{pu}\Rightarrow g_{ps}$), and (3) applying the progressive knowledge transfer operation (PKT). The comparison results are depicted in Tab.~\ref{table_CWT}. For the first variant, since the weights of the two networks are independent, the classifier $g_{pu}$ cannot benefit from the pseudo-supervised classifier $g_{ps}$, which affects the generation of pseudo supervision and limits the performance of the classifier $g_{ps}$, thus performing poorly. For the second variant, the classifier $g_{ps}$ may overcorrect the estimator bias, which limits the overall performance. In PKT, the performance of classifier $g_{pu}$ is improved by classifier $g_{ps}$, which in turn boosts the performance of classifier f through the data. The results show that PKT brings significant improvements in F1 score and AUC. 

\begin{table}[H]
    \caption{The results of the cooperation between classifier $g$ and $f$ on extreme imbalanced CIFAR-10}
    \vspace{-1mm}
    \label{table_CWT}
    \centering
    \resizebox{0.45\textwidth}{!}
    {$
        \begin{tabular}{ccccc}
        \toprule
        Setting & Classifier & ACC \%  & F1 \%& AUC \%\\
        \midrule
        \multirow{2}*{$g_{pu}\neq g_{ps}$} & $g_{pu}$ & 90.09$\pm$(0.17)& 14.48$\pm$(8.75)  & 68.73$\pm$(12.59) \\
        & $g_{ps}$ & 92.93$\pm$(0.53)& 47.90$\pm$(7.35)  & 86.71$\pm$(3.12) \\
        \midrule
        $g_{pu}\Rightarrow g_{ps}$  & $g_{ps}$ & \textbf{92.99$\pm$(1.24)} & 56.81$\pm$(3.39)  & 87.29$\pm$(2.03)\\
        \midrule
        \multirow{2}*{PKT} & $g_{pu}$ & 92.63$\pm$(0.41) & 47.77$\pm$(4.32) & 89.14$\pm$(2.21) \\
        & $g_{ps}$ & 89.72$\pm$(3.04) & \textbf{58.52$\pm$(4.89)} & \textbf{89.99$\pm$(0.88)} \\
        \bottomrule
        \end{tabular}
    $
    }
    \vspace{-0.1in}
\end{table}




\section{Conclusion}
In this paper, we propose a novel positive and unlabeled (PU) learning framework, named PSPU, which first leverages basic PU net with pseudo-supervised technique to handle the problem of overfitted risk estimation. In the experiment, PSPU gets significant performance gain over SOTA PU learning method on different benchmarks including image classification and industry anomaly detection, and achieves more stable and better performance on a more challenging setting (i.e., imbalanced division).

\vspace{1.5cm}

\noindent\textbf{Acknowledgement.} This work was supported in part by Shanghai Science and Technology Commission (21511101200), National Natural Science Foundation of China (No. 72192821).


{
\footnotesize
\itemsep=-3pt plus.2pt minus.2pt
\baselineskip=13pt plus.2pt minus.2pt

\bibliographystyle{IEEEbib}
\bibliography{refs}

\begin{thebibliography}{10}

\bibitem{chen2023precision}
Song Chen, Yongqin Qiu, Jingmao Li, Kan Fang, and Kuangnan Fang,
\newblock ``Precision marketing for financial industry using a pu-learning recommendation method,''
\newblock {\em Journal of Business Research}, vol. 160, pp. 113771, 2023.

\bibitem{de2022network}
Mariana~Caravanti de~Souza, Bruno~Magalh{\~a}es Nogueira, Rafael~Geraldeli Rossi, Ricardo~Marcondes Marcacini, Brucce~Neves Dos~Santos, and Solange~Oliveira Rezende,
\newblock ``A network-based positive and unlabeled learning approach for fake news detection,''
\newblock {\em Machine learning}, vol. 111, no. 10, pp. 3549--3592, 2022.

\bibitem{du2015convex}
Marthinus~Christoffel Du~Plessis, Gang Niu, and Masashi Sugiyama,
\newblock ``Convex formulation for learning from positive and unlabeled data,''
\newblock in {\em Proceedings of the 32nd International Conference on International Conference on Machine Learning}, 2015, pp. 1386--1394.

\bibitem{niu2016theoretical}
Gang Niu, Marthinus~Christoffel du~Plessis, Tomoya Sakai, Yao Ma, and Masashi Sugiyama,
\newblock ``Theoretical comparisons of positive-unlabeled learning against positive-negative learning,''
\newblock {\em Advances in neural information processing systems}, vol. 29, pp. 1199--1207, 2016.

\bibitem{du2014analysis}
Marthinus~C du~Plessis, Gang Niu, and Masashi Sugiyama,
\newblock ``Analysis of learning from positive and unlabeled data,''
\newblock in {\em Advances in Neural Information Processing Systems}, Z.~Ghahramani, M.~Welling, C.~Cortes, N.~Lawrence, and K.~Q. Weinberger, Eds., 2014, vol.~27.

\bibitem{kiryo2017positive}
Ryuichi Kiryo, Gang Niu, Marthinus C~du Plessis, and Masashi Sugiyama,
\newblock ``Positive-unlabeled learning with non-negative risk estimator,''
\newblock {\em Advances in neural information processing systems}, 2017.

\bibitem{bekker2020learning}
Jessa Bekker and Jesse Davis,
\newblock ``Learning from positive and unlabeled data: A survey,''
\newblock {\em Machine Learning}, vol. 109, pp. 719--760, 2020.

\bibitem{wang2021asymmetric}
Cong Wang, Jian Pu, Zhi Xu, and Junping Zhang,
\newblock ``Asymmetric loss for positive-unlabeled learning,''
\newblock in {\em 2021 IEEE International Conference on Multimedia and Expo (ICME)}. IEEE, 2021, pp. 1--6.

\bibitem{claesen2015a}
Marc Claesen, Frank De~Smet, Johan~A.K. Suykens, and Bart De~Moor,
\newblock ``A robust ensemble approach to learn from positive and unlabeled data using svm base models,''
\newblock {\em Neurocomputing}, vol. 160, pp. 73--84, Jul 2015.

\bibitem{he2020instancede}
Fengxiang He, Tongliang Liu, Geoffrey~I Webb, and Dacheng Tao,
\newblock ``Instance-dependent pu learning by bayesian optimal relabeling,'' 2020.

\bibitem{xu2022split}
Chengming Xu, Chen Liu, Siqian Yang, Yabiao Wang, Shijie Zhang, Lijie Jia, and Yanwei Fu,
\newblock ``Split-pu: Hardness-aware training strategy for positive-unlabeled learning,''
\newblock in {\em Proceedings of the 30th ACM International Conference on Multimedia}, 2022, pp. 2719--2729.

\bibitem{sakai2018semi}
Tomoya Sakai, Gang Niu, and Masashi Sugiyama,
\newblock ``Semi-supervised auc optimization based on positive-unlabeled learning,''
\newblock {\em Machine Learning}, vol. 107, pp. 767--794, 2018.

\bibitem{su2021positive}
Guangxin Su, Weitong Chen, and Miao Xu,
\newblock ``Positive-unlabeled learning from imbalanced data,''
\newblock in {\em Proceedings of the 30th International Joint Conference on Artificial Intelligence, Virtual Event}, 2021.

\bibitem{zhao2022dist}
Yunrui Zhao, Qianqian Xu, Yangbangyan Jiang, Peisong Wen, and Qingming Huang,
\newblock ``Dist-pu: Positive-unlabeled learning from a label distribution perspective,''
\newblock in {\em Proceedings of the IEEE/CVF Conference on Computer Vision and Pattern Recognition}, 2022, pp. 14461--14470.

\bibitem{yang2021aso}
Xiangli Yang, Zixing Song, Irwin King, and Zenglin Xu,
\newblock ``A survey on deep semi-supervised learning,''
\newblock {\em IEEE Transactions on Knowledge and Data Engineering}, 2022.

\bibitem{hamilton2017inductive}
William~L. Hamilton, Rex Ying, and Jure Leskovec,
\newblock ``Inductive representation learning on large graphs,''
\newblock in {\em Proceedings of the 31st International Conference on Neural Information Processing Systems}, 2017, pp. 1025--1035.

\bibitem{wang2021enaet}
Xiao Wang, Daisuke Kihara, Jiebo Luo, and Guo-Jun Qi,
\newblock ``Enaet: A self-trained framework for semi-supervised and supervised learning with ensemble transformations,''
\newblock {\em IEEE Transactions on Image Processing}, vol. 30, pp. 1639--1647, jan 2021.

\bibitem{imani2019semihd}
Mohsen Imani, Samuel Bosch, Mojan Javaheripi, Bita Rouhani, Xinyu Wu, Farinaz Koushanfar, and Tajana Rosing,
\newblock ``Semihd: Semi-supervised learning using hyperdimensional computing,''
\newblock in {\em 2019 IEEE/ACM International Conference on Computer-Aided Design}, 2019, pp. 1--8.

\bibitem{berthelot2019mixmatch}
David Berthelot, Nicholas Carlini, Ian Goodfellow, Nicolas Papernot, Avital Oliver, and Colin Raffel,
\newblock ``Mixmatch: A holistic approach to semi-supervised learning,''
\newblock {\em Advances in neural information processing systems}, 2019.

\bibitem{sohn2020fixmatch}
Kihyuk Sohn, David Berthelot, Nicholas Carlini, Zizhao Zhang, Han Zhang, Colin~A Raffel, Ekin~Dogus Cubuk, Alexey Kurakin, and Chun-Liang Li,
\newblock ``Fixmatch: Simplifying semi-supervised learning with consistency and confidence,''
\newblock {\em Advances in neural information processing systems}, vol. 33, pp. 596--608, 2020.

\bibitem{zhang2021flexmatch}
Bowen Zhang, Yidong Wang, Wenxin Hou, Hao Wu, Jindong Wang, Manabu Okumura, and Takahiro Shinozaki,
\newblock ``Flexmatch: Boosting semi-supervised learning with curriculum pseudo labeling,''
\newblock {\em Advances in Neural Information Processing Systems}, vol. 34, pp. 18408--18419, 2021.

\bibitem{bergmann2021mvtec}
Paul Bergmann, Kilian Batzner, Michael Fauser, David Sattlegger, and Carsten Steger,
\newblock ``The mvtec anomaly detection dataset: a comprehensive real-world dataset for unsupervised anomaly detection,''
\newblock {\em International Journal of Computer Vision}, vol. 129, no. 4, pp. 1038--1059, 2021.

\bibitem{berthelot2019remixmatch}
David Berthelot, Nicholas Carlini, Ekin~D Cubuk, Alex Kurakin, Kihyuk Sohn, Han Zhang, and Colin Raffel,
\newblock ``Remixmatch: Semi-supervised learning with distribution alignment and augmentation anchoring,'' 2019.

\bibitem{wang2023beyond}
Xinrui Wang, Wenhai Wan, Chuanxin Geng, Shaoyuan LI, and Songcan Chen,
\newblock ``Beyond myopia: Learning from positive and unlabeled data through holistic predictive trends,''
\newblock {\em arXiv preprint arXiv:2310.04078}, 2023.

\bibitem{chen2019variational}
Hui Chen, Fangqing Liu, Yin Wang, Liyue Zhao, and Hao Wu,
\newblock ``A variational approach for learning from positive and unlabeled data,''
\newblock {\em arXiv preprint arXiv:1906.00642}, 2019.

\bibitem{chen2020self}
Xuxi Chen, Wuyang Chen, Tianlong Chen, Ye~Yuan, Chen Gong, Kewei Chen, and Zhangyang Wang,
\newblock ``Self-pu: Self boosted and calibrated positive-unlabeled training,''
\newblock in {\em International Conference on Machine Learning}, 2020, pp. 1510--1519.

\end{thebibliography}
}

\end{document}